\documentclass[conference]{IEEEtran}
\usepackage{cite}
\usepackage{algorithm}
\usepackage{algpseudocode}
\usepackage{url}
\usepackage{hyperref}
\usepackage{amsmath,amssymb,amsfonts}
\usepackage{graphicx}
\usepackage{textcomp}
\usepackage{xcolor}
\def\BibTeX{{\rm B\kern-.05em{\sc i\kern-.025em b}\kern-.08em
    T\kern-.1667em\lower.7ex\hbox{E}\kern-.125emX}}
    
\begin{document}

\title{AI based approach to Trailer Generation for Online Educational Courses}
\author{\IEEEauthorblockN{1\textsuperscript{st} Prakhar Mishra}
\IEEEauthorblockA{\textit{IIIT}\\
Bangalore, India \\
prakhar.mishra@iiitb.ac.in}
\and
\IEEEauthorblockN{2\textsuperscript{nd} Chaitali Diwan}
\IEEEauthorblockA{\textit{IIIT}\\
Bangalore, India \\
chaitali.diwan@iiitb.ac.in}
\and
\IEEEauthorblockN{3\textsuperscript{rd} Srinath Srinivasa}
\IEEEauthorblockA{\textit{IIIT}\\
Bangalore, India \\
sri@iiitb.ac.in}
\and
\IEEEauthorblockN{4\textsuperscript{th} G. Srinivasaraghavan}
\IEEEauthorblockA{\textit{IIIT}\\
Bangalore, India \\
gsr@iiitb.ac.in}}

\maketitle

\begin{abstract}
In this paper, we propose an AI based approach to Trailer Generation in the form of short videos for online educational courses. Trailers give an overview of the course to the learners and help them make an informed choice about the courses they want to learn. It also helps to generate curiosity and interest among the learners and encourages them to pursue a course. While it is possible to manually generate the trailers, it requires extensive human efforts and skills over a broad spectrum of design, span selection, video editing, domain knowledge, etc., thus making it time-consuming and expensive, especially in an academic setting. The framework we propose in this work is a template based method for video trailer generation, where most of the textual content of the trailer is auto-generated and the trailer video is automatically generated, by leveraging Machine Learning and Natural Language Processing techniques. The proposed trailer is in the form of a timeline consisting of various fragments created by selecting, para-phrasing or generating content using various proposed techniques. The fragments are further enhanced by adding voice-over text, subtitles, animations, etc., to create a holistic experience. Finally, we perform user evaluation with 63 human evaluators for evaluating the trailers generated by our system and the results obtained were encouraging.
\end{abstract}


\section{Introduction}
The growth of the internet has significantly increased the amount of free instructional content. These resources are offered not only by big institutions but also by individual content creators over various platforms such as Coursera, Udemy, YouTube, etc. This increase in content production rate has resulted in the creation of redundant courses and tutoring videos for many topics over time. In spite of advantages like on-demand accessibility, the abundance of options has increased confusion and made it more challenging to select a course that might be in line with learner's interests. And often, enrolling to a course that doesn't meet the learner's expectations for a course's curriculum and other aspects such as expected level of commitment, the availability of support, etc., causes the learner to lose motivation and eventually drop the course.~\cite{simpson2013student,hartnett2011examining}. 

This problem can be tackled to a certain extent by presenting a video trailer to the learners before the start of the course (learning pathway) to help them quickly glance through the pathway and get an overall idea of the course content and its format~\cite{gayoung2016study,wong2016factors,stacey2014pedagogy}.

The idea of \emph{Trailers} is not brand-new, and the film industry has been using them extensively for a while. Trailers, in context of movies are mostly about advertising. They notify viewers about an upcoming movie while generating interest among them.
Often the effectiveness of a trailer affects the perception of the movie, even before it is released publicly. 
The course trailers serve a greater purpose in the educational context than simple course promotion. Before beginning the learning journey, they aid in helping learners set realistic expectations for their learning outcomes and competency mastery.

Concept of trailers might resemble with that of summarization~\cite{zhang2020pegasus,raffel2019exploring,mihalcea2004textrank}, but apart from incorporating a few elements of summarization like shortening and abstracting out information from substantial sized input source, trailers are different in terms of their motivation, purpose and the impact they create on the end users. Unlike summaries, trailers need not be complete in their coverage. 
Also, they are designed to give glimpses of a few interesting segments of the narrative without revealing the main plot or climax of the underlying narrative~\cite{lienhart1997video}.
Although there is no clear demarcation of what a climax is in academic narratives, based on our analysis of many academic course trailers in popular MOOCs (Massive Open Online Courses) such as Udemy\footnote{\url{https://www.udemy.com}} and Coursera\footnote{\url{https://www.coursera.org}}, we see prevalence of a common pattern in trailer timelines. The timeline starts with an introduction about the course and the instructor and ends with a call-to-action (CTA) which offers opportunity to the learners to take action or start the course. In between, there are several elements and factoids about the course and its contents, that aim to arouse viewer interest. 

The current approach of generating trailers is manual, cumbersome and time-consuming, it requires someone with relevant skills like designing, video editing, and a subject matter expert to help in curating the trailer content.
Although, there are software products like Apple iMovie\footnote{\url{https://www.apple.com/in/imovie}}, Windows Movie Maker\footnote{\url{https://www.microsoft.com/en-us/p/movie-maker-video-editor/9mvfq4lmz6c9}} and others that people can use for generating trailers by performing basic editing like cuts, merging frames, etc. Yet the content to be placed in the trailer has to be curated entirely by a human expert. 

\begin{figure}
    \centering
    \includegraphics[width=0.4\textwidth]{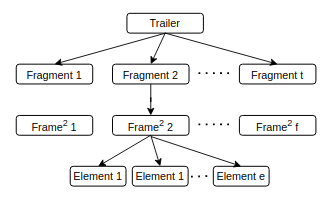}
    \caption{Trailer Structure}
    \label{fig:trailer_structure}
\end{figure}

In our work, we propose a semi-automatic template based framework for generating video trailers for learning pathways, which are a sequence of related educational documents of various forms~\cite{diwan2019automatic,chi2009ontology,shmelev2015approach}. Here, most of the content that is placed in the trailer is auto-generated with a scope for taking inputs from the creator. The framework for trailer generation consists of various essential trailer fragments arranged as a timeline of the trailer. Each fragment is composed of a sequence of frames that are coherent within themselves in terms of the topical information they present. And inherently, each frame is composed of various types of elements and their properties like font size, text styling, image size, etc. Fig.~\ref{fig:trailer_structure} shows the illustration for the same.

\begin{figure*}
    \centering
    \includegraphics[width=\textwidth]{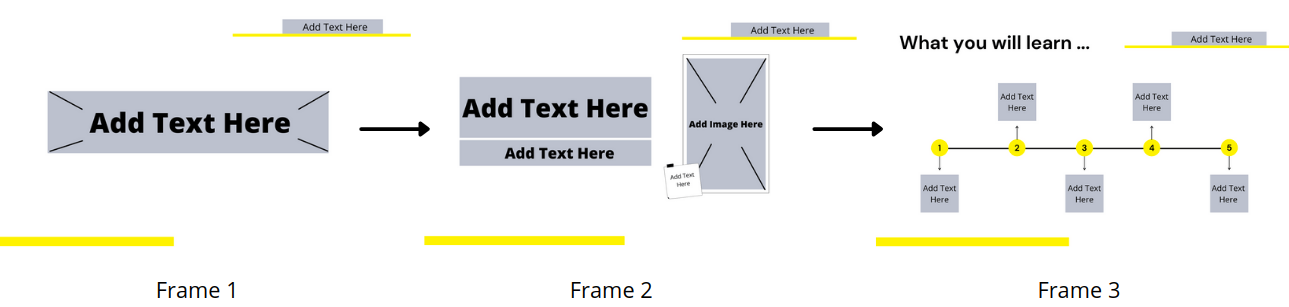}
    \caption{Illustration of Frames}
    \label{fig:trailertemplate}
\end{figure*}

Once all the elements are generated and placed at their respective positions within a frame of a trailer fragment, a template is applied to it. The template consists of the multi-modal experiences such as voice-over, subtitles, sounds, animations, etc. It also determines the elements of the trailer design such as number and ordering of fragments, frames and elements. 
Fig.~\ref{fig:trailertemplate} shows the visual view of some of the frames for one of the templates with it's corresponding elements and their positioning in the frames.

\section{Related Work}
\label{sec:relatedwork}
There are studies that discuss the idea, use and motivation of having trailers for academic courses~\cite{gayoung2016study,wong2016factors,stacey2014pedagogy}. Also, there are online educational platforms like Coursera and Udemy which have course trailers. However, we could not find literature on approaches to generating trailers for academic courses. Hence, in the following paragraphs we discuss some of the pioneering works of trailer generation in-general across other domains. Trailer generation can also be seen as special case of larger research interest of adding an element of surprise to the engage receiver’s attention in midst of information overload~\cite{varshney2013surprise,varshney2019must}.

Authors in~\cite{brachmann2015automatic,hermes2006automatic,irie2010automatic,smith2017harnessing} present an approach for automatic trailer generation from movies as input. Hermes et al.~\cite{hermes2006automatic} create trailers for action movies by analyzing audio and video signals present in movies and automatically detecting features like faces, scene cuts, sound-volume, etc and use ontology of the corresponding domain for producing trailers. Irie et al.~\cite{irie2010automatic} propose a movie trailer generation method which extracts symbols like title logo, main theme music and selects impressive shot or speech segments based on clustering methods and EM algorithm. Brachmann et al.~\cite{brachmann2015automatic} propose an approach of generating action movie trailers using the concept of trailer grammar, knowledge base and various ML techniques for analyzing audio and images present in the movie. Smith et al.~\cite{smith2017harnessing} propose a system that understands and encodes the patterns and emotions present in horror movies using Convolution Neural Networks(CNN). 

All the above methods use visual and audio cues to derive the trailer frames, whereas we use raw text data and build the necessary discriminative and generative Neural Network models to create frames and its elements to be placed in the trailer.  

Hesham et al. in~\cite{hesham2018smart} explore the idea of creating movie trailers from their subtitles. They first classify the movie by genre, identify important keywords and then rank important subtitles. The trailer is then generated by stacking the movie time-frames corresponding to the important subtitles. 
Gaikwad et al. in~\cite{gaikwad2021plots} propose a technique to create previews of movies by utilizing subtitles and finding the most representative scenes by matching them with the plot summaries. Chi et al.~\cite{chi2020automatic} propose an approach to automatically create marketing-style short videos for a given product page url by extracting elements and their styles present in the product html page under specified tags. 

Unlike the aforementioned works which primarily focus on generating trailers based on an extractive strategies, in our work we develop various modules that comprehend input document and generate content for the trailer either by paraphrasing or by using Natural Language Generator based model. 

As far as we know, automatic/semi-automatic generation of video trailers for learning pathways is unexplored. Our proposed approach of video trailer generation using Machine Learning, Natural Language Processing and Generation techniques is also unique.

\section{Proposed System}
\label{sec:proposedsystem}
We propose a framework for trailer generation consisting of different trailer fragments that form a trailer timeline, generation of the trailer fragments and finally applying templates that determine the look and feel of the trailer.
Based on our analysis of multiple trailers presented for various online courses offered on various educational platforms like Coursera and Udemy, we designed and structured our trailer elements, fragments and overall flow of the trailer. 

We propose a trailer timeline consisting of 7 trailer fragments namely,
\begin{figure*}
    \centering
    \includegraphics[width=0.6\textwidth]{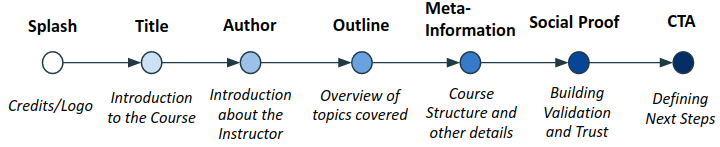}
    \caption{Trailer Timeline}
    \label{fig:trailertimeline}
\end{figure*}
Splash, Trailer Title, Author Details, Outline, Meta-Information, Social Proof and finally the Call-to-Action. Figure~\ref{fig:trailertimeline} shows the timeline of all the above-mentioned fragments in the trailer.
Each of these fragments define a specific part of the trailer, their purpose and their importance in the trailer. We define the fragments in detail in further part of this section. As discussed earlier, fragments are composed of a sequence of frames and each frame is composed of various types of elements and their properties.

\begin{figure*}
    \centering
    \includegraphics[width=\textwidth]{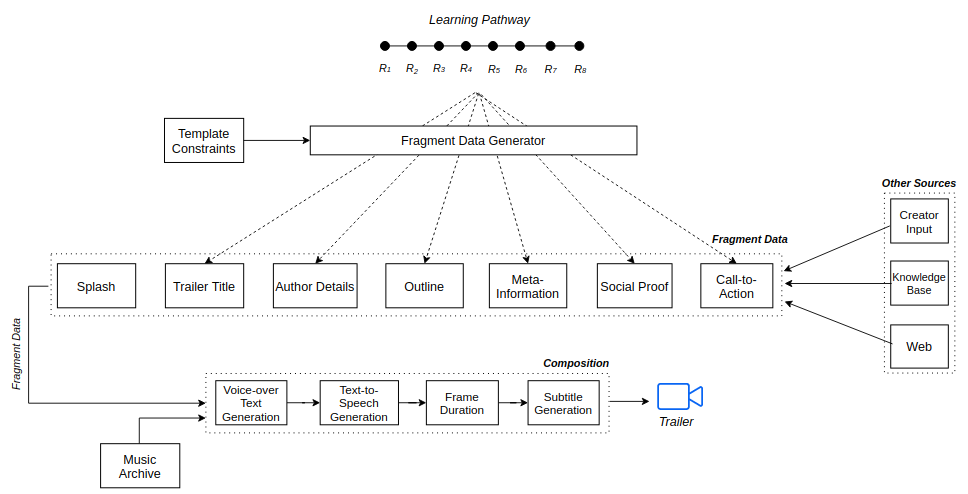}
    \caption{Trailer Generation Flow}
    \label{fig:outlinetrailer}
\end{figure*}

The overall approach for trailer generation is illustrated in Fig.~\ref{fig:outlinetrailer}. 
All the resources mapped to a learning pathway form the input to our \emph{Fragment Data Generator (FDG)} module.
Template constraints that define the elements, fragments and frames also form the input to \emph{FDG}. 
The \emph{FDG} along with other sources like creator's input, any images or information from the web or knowledge bases, etc., can be incorporated into the frames or the fragments. 
Once the elements for all the frames across all the fragments are generated, we pass it to the composition module for adding in other important aspects of the trailer like voice-over, subtitles, sounds, etc., to add to its multi-modal experience. 

\subsection{Fragment Data Generation}
\label{themegeneration}
\begin{figure*}
    \centering
    \includegraphics[width=0.7\textwidth]{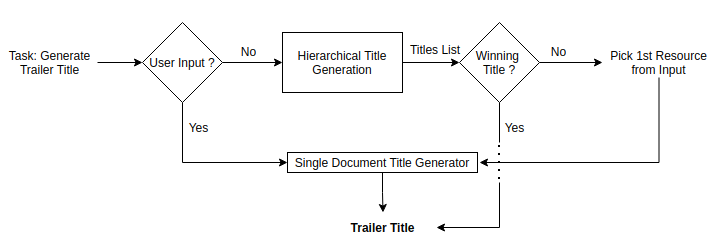}
    \caption{Trailer Title Fragment Generation Flow}
    \label{fig:trailer_title}
\end{figure*}

Following are the proposed trailer fragments arranged in the order of their appearance in the trailer timeline-
\paragraph*{Splash Fragment} The idea of splash fragment is to display any introductory information related to the trailer such as credits, software logo, etc., mostly obtained from creator's input. This optional fragment could also be the last fragment in the trailer depending on the creator's preference.
\paragraph*{Trailer Title Fragment}
In this fragment we generate a short yet representative title for the entire trailer, hence giving a quick idea about the topic that summarizes the underlying pathway or the set of resources. We apply \emph{Hierarchical Title Generation} model~\cite{mishra2021automatictitle} over the resources mapped to the learning pathway to get the list of trailer titles. We select a title among them based on their Term Frequency. In case, none of the titles are above a threshold, we fall back on the fact that the first resource in the pathway is the proxy to the introductory resource, and we generate the trailer title for it by applying \emph{Single Document Title Generator}~\cite{mishra2021automatic,tan2017neural}. Figure~\ref{fig:trailer_title} shows the trailer title fragment generation flow.

\paragraph*{Author Details Fragment}
A quick introduction about the author or the instructor of the learning pathway could help the learners build an implicit connect and trust. 
Majority of the elements in the \emph{Author Details Fragment} like author names, affiliations and author's image are expected from the creator while creating the trailer. 
Template constraints such as addressing multiple authors with different frame elements, handling and getting relevant images to be put in this fragment etc are also obtained from trailer creator. These inputs and template constraints are plugged in the automation system to fill the overall author frame. 
Additionally, we crawl the web to get relevant images, for example: we crawl the web and get relevant affiliation images and place it in the desired coordinates as defined by the template. Also for the templates that allow for having only the frontal face of author, we make use of an open-sourced face recognition model\footnote{\url{https://docs.opencv.org/3.4/db/d28/tutorial_cascade_classifier.html}} to crop the face from the uploaded author image. 
In case no author image is provided to the system by the creator, we place a dummy caricatured relevant sized image. Similarly, we have defined defaults for the features, frames and templates in case there is no input from the trailer creator. For example, when multiple authors exists, we display information w.r.t to the the first author entered by the creator and treat him/her as the primary instructor and rest all the authors are abstracted by placing them under the ``and others" category.

\paragraph*{Outline Fragment} 
This fragment gives an idea about the specific topics that would be covered in the learning pathway. 
This could help in setting learners' expectation in terms of the topics covered and in deciding whether the content aligns to his/her end goals. 
For this we use~\emph{Single Document Title Generator}~\cite{mishra2021automatic,tan2017neural} model to generate titles for all the resources in the learning pathway which represents the outline of the learning pathway.

\begin{figure*}
    \includegraphics[width=0.95\textwidth]{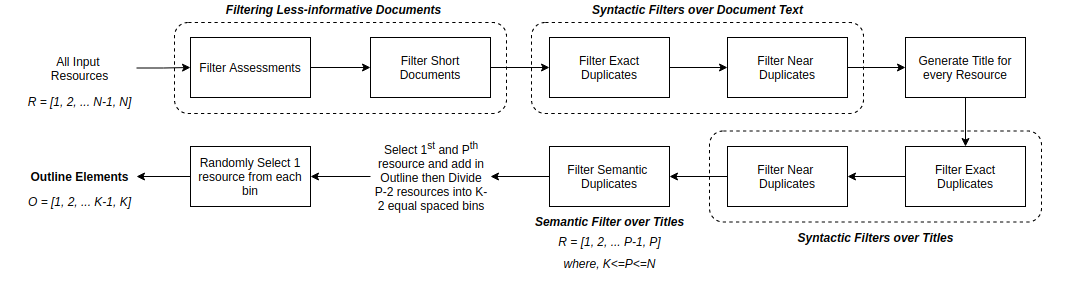}
    \caption{Outline Text Selection}
    \label{fig:outlinetextselection}
\end{figure*}

Every template under the outline fragment limits the number of text elements to be listed on the screen with the aim to balance aesthetics and information at the same time. To adhere to this prior constraint, we design a multi-step process to select diverse, yet impactful set of elements from a relatively larger list of outlines generated in the previous step. Fig.~\ref{fig:outlinetextselection} shows the entire pipeline of Outline Text Selection.

Let $K$ be the number of text elements that the frame requires and $N$ be the total number of resources we have as input and let $K<N$. We start with all the resources (N) given by the user and remove any instance of assessments and short documents under the assumption that such documents won't hold much informational content. After this we remove any occurrence of exact duplicates and near duplicates in the remaining set and pass the remaining resource list to the title generator system to generate title for every resource. 

Post this, we fix the first and the last position of the outline with the first and last resource title.
We specifically do this action because of the inherent ordering present in the input resource as a part of learning pathway. Also intuitively, picking first and last sets a bound over the topic space to be covered under a particular course. 

Finally on this reduced set, we divide the space into bins of equal size from which we randomly sample one outline element from each bin to remaining $K-2$ positions in the outline list.
We use threshold based Jaccard and cosine similarity for filtering syntactic and semantic duplicates respectively. 
The Jaccard similarity between any two documents is calculated as an intersection over union of word sets for both documents. It helps us get sense of syntactic similarity between documents. For calculating cosine similarity, we vectorise our inputs using pre-trained Sentence Transformers~\cite{reimers2019sentence} and then measure the semantic closeness between them using cosine similarity.

\begin{algorithm} 
	\caption{Duplicates Filter}
	\begin{algorithmic}[1]
	   \State $resources=Array(1,2,\ldots,N-1,N)$
	   \State $remaining\_resources=Array(1, N)$
	  
		\For {$i \gets 2$ to $N-1$}    
		    \State $scores=Array()$
		    \For {$r \gets remaining\_resources$}
    		    \State $scores \gets calculate\_similarity(i, r)$
    		\EndFor
    		
			\If{$max(scores) < threshold$}
			    \State $remaining\_resources \gets i$
			\EndIf
		\EndFor
	\State return $remaining\_resources$
	\end{algorithmic} 
	\label{alg:filter}
\end{algorithm}
Since every pathway is composed of different resources of various properties like length, style, etc., having one threshold that fits all does not work. Hence, our threshold is adaptable in a way that guarantees at-least $K$ items are selected post any of the syntactic or semantic pruning steps. The threshold search space is between 0 to 1 where for efficiency and tractability we quantize it at 0.1. Then for each threshold we get remaining resources as defined in Algorithm~\ref{alg:filter}. Finally the threshold that guarantees at-least $K$ items and possibly reduces the input set by maximum is chosen as the final threshold.

\paragraph*{Meta-Information Fragment}
The idea of having Meta-Information Fragment is to inform learners about other important aspects of the course like course structure, total reading time, total number of resources, etc.  We believe this would help learners understand more about the learning pathway or resources apart from just knowing the topics that would be covered. Also, such information can be used by learners in charting out their learning hours and estimating the efforts it would take for the successful completion of the course. Some of the elements that we generate automatically as part of this fragment are: generating topical word clouds~\footnote{\url{https://pypi.org/project/wordcloud/}} bases on word frequencies after pre-processing like stop-word removal, estimating total reading time based on average reading speed statistics and other pathway level derived statistics like total resources, availability of discussion forum, etc.

\paragraph*{Social Proof Fragment}
Social Proof is one of the most prominent ways of social influence and is based on the heuristic that the users follow others similar to them when uncertain~\cite{cialdini2009influence}. We collect these statistics from the deployed learning environments. This information is added to the video trailer over time when different learners take this course and the analytical data is available.

\paragraph*{Call-to-Action Fragment}
CTA is a marketing term which is designed to push the audience in taking the desired actions. It is an important aspect of any trailer because all of the enthusiasm that is built in a learner while watching the trailer is of no use if the learner is not clear on the next actionable~\cite{CTA1,CTA2} item. In our system, we randomly select phrases from a set of pre-defined list of potential key-phrases to be placed on the screen at a pre-defined location under this fragment. Some of the phrases we use are `Start your learning today', `Let's get started', `Are you ready?', etc., along with the action that will take the learner on the learning pathway.

\subsection{Additional Elements}
In this subsection, we discuss two other interesting elements that we propose to be added to the trailers, namely, \emph{Definition Extractor} and \emph{Paraphraser}. These are shown as suggestions to the trailer creator and it's up to the creator to include them and decide their placement in the trailer.
\paragraph*{Definition Extractor}  
Definitions are descriptive elements that we believe can help in introduction of concepts. To select the definition from the learning resource, we propose a discriminative model that classifies a given piece of text into Definition or Non-Definition class. For building the classifier model, we use a dataset\footnote{\url{http://nlp.uniroma1.it/wcl/}} that contains positive and negative definition candidates extracted from Wikipedia for various topics. Our best performing model is a fine-tuned DistilBERT-base-uncased\footnote{\url{https://huggingface.co/distilbert-base-uncased}} model with a Definition class F1-score of 0.96 and Non-Definition class F1-score of 0.97 on the test set.

\paragraph*{Paraphraser} 
We believe that this is an useful utility that can be used in the Outline and Trailer title fragments. This gives the creator an ability to re-write concisely any substantially larger textual content present in any frame. We use a publicly available pre-trained model\footnote{\url{https://github.com/ramsrigouthamg/Questgen.ai}} for this task which fine-tunes a large sized T5 (Text-to-Text Transfer Transformer)~\cite{raffel2019exploring} model on a parallel corpus of sentence and it's corresponding paraphrase. 

\subsection{Video Composition}
\label{sec:videocomposition}
Video Composition module is responsible for stitching together all the elements that need to be part of the trailer, such as the Frame data, Voice-over text, Text-to-Speech (TTS), etc., into a trailer video. Fig.~\ref{fig:outlinetrailer} pictorially shows the overall flow of the various components that are a part of the video composition. We use Python's MoviePy library\footnote{~\url{https://zulko.github.io/moviepy}} as our choice for video editing and composition of the templates as it provides us with all the necessary editing functions like inserting text, concatenations and cuts, which we use to draft our templates.

\begin{figure*}
    \centering
    \includegraphics[width=0.9\textwidth]{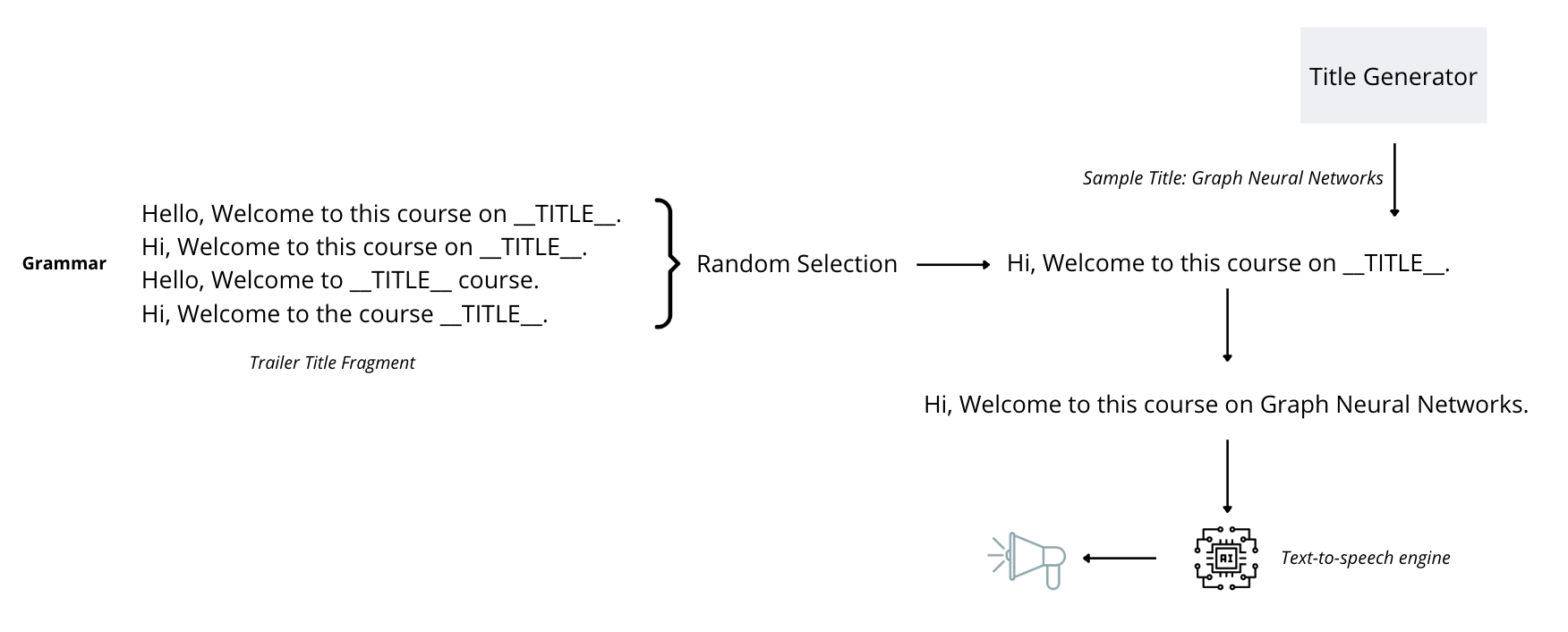}
    \caption{Flow of Grammar selection to Voice-over generation}
    \label{fig:textgrammar}
\end{figure*}

After the frame-level data elements are in-place, the next step is to generate voice-over text for each of the frames. Voice-over text is defined as the spoken-text that the narrator speaks while a frame is displayed on the screen. 
For this, we select grammar from a pre-defined set of slot based text grammars which we define per frame. The slots in the grammar are nothing but the screen's text elements. 
Finally, once the Voice-over Text is generated for every frame, we pass them through the IBM Watson's Text-to-speech (TTS) API\footnote{\url{https://cloud.ibm.com/catalog/services/speech-to-text}} with relevant parameters such as voice-type, gender, etc., by choosing from a list of speaker profiles to get the audio files for every frame. 
Fig.~\ref{fig:textgrammar} illustrates the flow from grammar selection to voice generation for the Trailer Title Fragment.
We then derive the frame duration accordingly to make sure that the visual and audio aspects of the frames are in sync and minimize any kind of lag on either ends. Finally, along with all the above details, we input template constraints like positioning of elements, and styles, user preferences, and some basic animations like fade-in and fade-out settings to come up with the final trailer.

\section{Experiments}
\label{sec:expandresults}
In this section, we describe the dataset, evaluation strategy and results obtained for the trailers generated by our proposed system.

\paragraph*{Dataset} 
Apart from the datasets which we have used for training and evaluating specific modules that are responsible for generating fragment relevant data. We created three different learning pathways for our experiments and evaluation of the generated trailers. 
Each learning pathway differs with each other in the number of resources and stylometry. 
Two of the pathways are based on text book chapters with difference in number of resources mapped, and one pathway is video lectures. We tried to take different pathways to evaluate our model's flexibility on different types of learning pathways.
First one was created by sampling some chapters sequentially from a freely available Machine Learning textbook~\cite{gareth2013introduction}. For second, we chose the speech-to-text transcription of a week's video lectures from an academic course on NLP. Our third learning pathway is the entire ML textbook~\cite{gareth2013introduction}\footnote{Datasets can be found at: \url{https://bit.ly/3ro3JLO}}. 
All the three corpus are analogous to learning pathways as they are all semantically coherent, progressive and share the same global topic. 

\paragraph*{Evaluation and Results}
Trailers can be seen as generative tasks with an inherent notion of creativity. 
Here the objective evaluation is not straight-forward because the effectiveness of a trailer is highly subjective and relies on the human perception. 
However, we think that human evaluation on various trailers generated can give us a good perspective on the quality of the trailers. 
We had 63 human evaluators consisting of Engineering graduates, Post-graduates and PhD students well versed in the technical domain that represent our dataset. 

We evaluate 6 trailers\footnote{Sample Trailers: \url{https://bit.ly/3Hscie9}} in total that were generated from 3 different learning pathways as discussed above, i.e., 2 trailer per learning pathway. These two trailers are based on two templates T1, T2 created by us. Both the templates differ in aesthetics and level-of-detail(LOD). The evaluation for each trailer was done on a set of 8 questions on Likert-scale from 1 to 5, where 1 would mean very poor and 5 would mean very good. 

\begin{table}
    \centering
    \begin{tabular}{|p{.02\linewidth}|p{.8\linewidth}|}
    \hline
         1 & The first trailer looked more catchy compared to the second one. Being generated by an AI agent, both seems to be good. \\
         2 & Looks amazing. Great work! \\
         3 & You guys have truly done a remarkable work! \\
         4 & Good job, keep it up! \\
         5 & Great! \\
    \hline
    \end{tabular}
    \caption{Positive comments}
    \label{tab:poscomments}
\end{table}
\begin{table}
    \centering
    \begin{tabular}{|p{.02\linewidth}|p{.8\linewidth}|}
    \hline
         1 & Maybe I just felt that he was conveying info too fast \\
         2 & As of now, it sounds a bit robotic. Some improvements w.r.t the TTS can help make it better. \\
         3 & Slowing the video when the information that is being conveyed is relatively dense would be helpful. For example, when going through the list of topics, speaking slowly helps. When giving instructor names, one can be fast. \\
         4 & Also, if there's some way to bring viewer's attention to the part of the slide that's being mentioned, that would be better where the content is not sequential. \\
         5 & Remove the date from the frame. Add something about what can they do once they learn the course(what type of problems can they solve) \\
    \hline
    \end{tabular}
    \caption{Improvements suggested by users}
    \label{tab:negcomments}
\end{table}

There were three separate groups of evaluators. 
Each group was provided with 2 trailers based on 2 templates for the same pathway. We thoughtfully perform this diversification to simulate the cluster sampling procedure, since showing all 6 trailers to the same evaluators would have created boredom, resulting in not so accurate evaluation.

\begin{figure*}
    \includegraphics[width=1\textwidth]{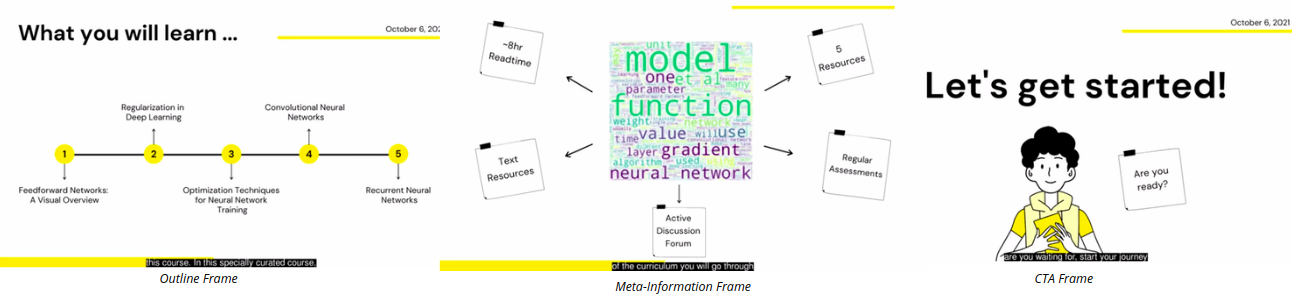}
    \caption{Trailer Fragments}
    \label{fig:TrailerFragments}
\end{figure*}

We also encouraged the evaluators to give free comments for the trailers they evaluated, as this would help us improve our system in future iterations. Table. ~\ref{tab:poscomments} and ~\ref{tab:negcomments} lists down some of the positive comments and improvements suggested by the users. Fig.~\ref{fig:TrailerFragments} shows some of the trailer fragments generated by our proposed system\footnote{Detailed demo walk-through: \url{https://www.youtube.com/watch?v=06VVuAlFhTk}}. 

Following is the list of 8 questions that were asked to the evaluator during the evaluation. 
The text in \emph{italics} highlights the broader aspect of the evaluation feature.
\begin{itemize}
  \item[] Q1. Did you find the trailer to be \emph{self-contained}?
  \item[] Q2. How were the \emph{fonts and styles} used in the trailer in terms of readability?
  \item[] Q3. How did you find the \emph{length and pace} of the trailer?
  \item[] Q4. As a user, how \emph{impressed} are you with this trailer overall?
  \item[] Q5. Could this trailer \emph{evoke interest} in someone taking this course? (Ignoring any prior inclination to the topic)
  \item[] Q6. How was the \emph{average duration} of each frame?
  \item[] Q7. Based on the trailer you just saw, do you think you have a good \emph{impression} of the course now?
  \item[] Q8. How did you find the \emph{sync between the audio and visuals} you saw?
\end{itemize}

\begin{figure*}
    \centering
    \includegraphics[width=0.5\textwidth]{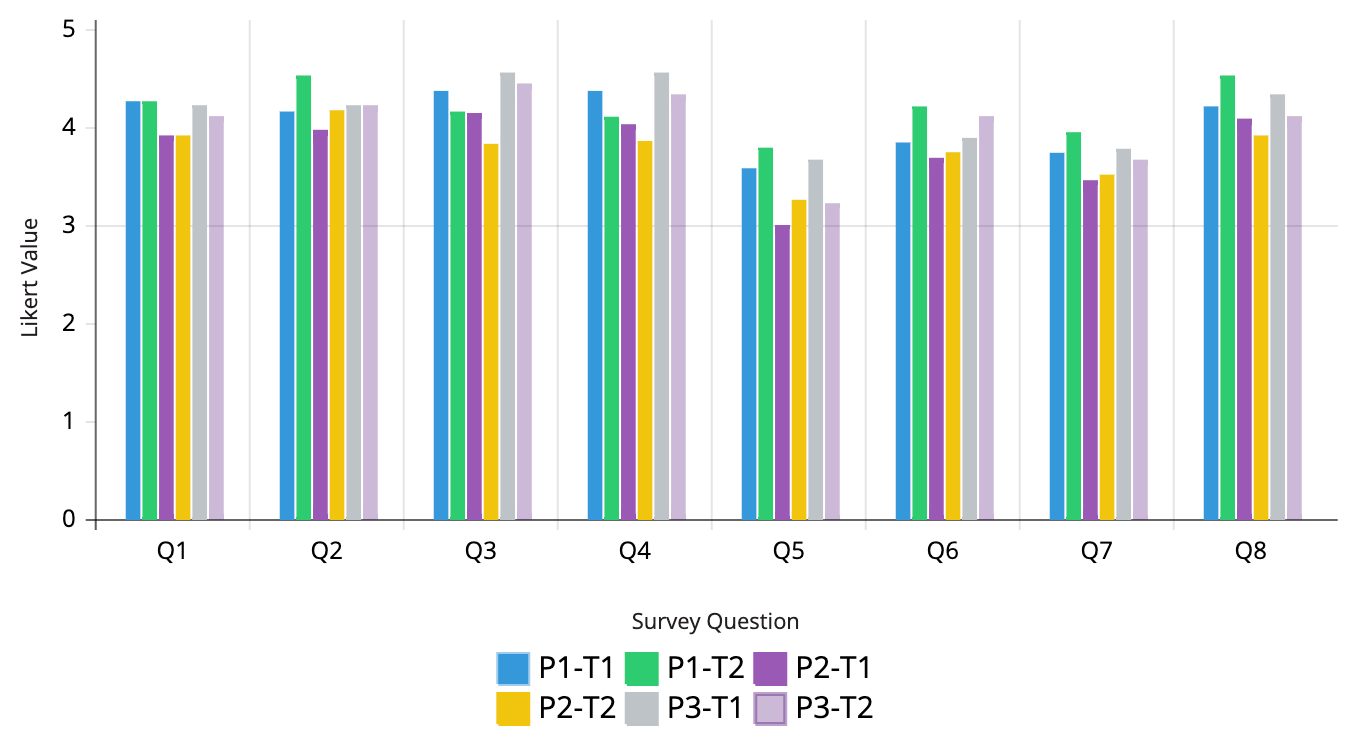}
    \caption{Average scores per Survey Question for all 3 pathways and trailers. Here P1, P2, P3 represent 3 Pathways and T1, T2 represent Templates}
    \label{fig:likert}
\end{figure*}
As can be seen in Fig.~\ref{fig:likert}, the scores obtained for each of the survey questions are good and far above the average(score of 3) for almost all the trailers generated by our approach. 
Also, in our study, we found both the templates performed equally good.
However, for Q5, the average scores is relatively lower compared to other questions. On digging deeper we found some of the comments of total 24 comments we received mentioned about the difficulty of the course for not getting interested in the course. This could mean that this question (Q5) is more subjective.

\section{Conclusions and Future Work}
\label{sec:conclusionandfutureword}
In this paper, we presented a novel framework for automatically generating video trailers for a learning pathway using ML and NLP techniques.
We validated our trailers on multiple corpus of varied granularity with human evaluation and the results obtained were encouraging. This approach can be adapted to different domains given enough data to train the models involved in the entire process. We believe that this approach can lay foundation to building more advanced versions of trailer.

In future, we plan to improve the existing system by incorporating suggestions obtained in the user evaluation and adding more interesting themes like automatically detecting learning outcomes given the resources. 
We also intend to create an interactive dashboard to take inputs from the creator and allow the creator to make edits to the auto-generated content.

\section*{Acknowledgment}
We thank the Center of Excellence on Cognitive Computing, funded by Mphasis F1 Foundation for funding this research. We also thank Dr. Prasad Ram and Gooru team (https://gooru.org) for the topical discussions and encouragement. 

\bibliographystyle{IEEEtran}
\bibliography{IEEEabrv,trailer}

\end{document}